# Fraud/Uncollectible Debt Detection Using a Bayesian Network Based Learning System:
## A Rare Binary Outcome with Mixed Data Structures

**Kazuo J. Ezawa**  
Room 7E-523  
kje@ulysses.att.com  

**Til Schuermann**  
Room 7E-530  
til@ ulysses.att.com  

AT&T Bell Laboratories  
600 Mountain Avenue  
Murray Hill, N. J. 07974

## Abstract

The fraud/uncollectible debt[1] problem in the telecommunications industry presents two technical challenges: the detection and the treatment of the account given the detection. In this paper, we focus on the first problem of detection using Bayesian network models, and we briefly discuss the application of a normative expert system for the treatment at the end. We apply Bayesian network models to the problem of fraud/uncollectible debt detection for telecommunication services. In addition to being quite successful at predicting rare event outcomes, it is able to handle a mixture of categorical and continuous data. We present a performance comparison using linear and non-linear discriminant analysis, classification and regression trees, and Bayesian network models.

## 1 INTRODUCTION

Every year, the telecommunications industry incurs several billion dollars in uncollectible revenue, the vast majority of which is in the form of net bad debt (NBD).[1] In this competitive industry, cost efficiency is paramount for success and survival, and NBD is a prime target for high-impact cost cutting. We seek to provide the business with two tools: one which will surpass existing detection methods, and a second which will provide recommendations of how to handle the customers identified by the first. In this paper we present the Advanced Pattern Recognition and Identification (APRI) system, which is a Bayesian network supervised machine learning system, and compare its performance to existing methods such as statistical discriminant analysis and classification and regression trees (CART) [Breiman, Friedman, Olshen and Stone, 1984].

As is no surprise, the NBD detection problem is a difficult one. We are essentially trying to find a small but valuable needle in a rather large haystack. For one, we are dealing with a binary outcome (a call or a customer is either "good" or "bad" from a collections perspective), where the outcome of interest occurs quite rarely: 1 - 2 % of the time, in fact.[2] Second, falsely classifying a good customer as bad could have dire consequences: we could lose such a customer. APRI allows the detection and identification of potential candidates: how to handle them is another matter. For this we are developing Normative Expert System Development Tool (NESDT) to provide intelligent decision support for account treatment recommendation for such a candidate.

Finally, modeling is made still more difficult because the data at our disposal is both discrete and continuous in nature. Examples would be the total monthly bill of a customer (in dollars, which is continuous) and the area code of residence (which is categorical or discrete). There are methods designed to handle either one or the other kind of data effectively, but none deal with a mixed situation satisfactorily. In addition to the data characteristics which make modeling difficult, the sheer *size* of the data sets under consideration, while large by research standards (4-6 million records and 600-800 million bytes), are *small* by telecommunications industry standards. Some learning methods simply cannot hope to process this much data in a timely manner.

We will be comparing our APRI system to discriminant analysis and CART[3]. In addition we attempted a comparison with the classification tree method of C4.5 [Quinlan, 1993] but were unable to complete a run. The C4.5 implementation is simply unable to handle data sets as large as ours. However, we have an implementation of CART that is able to handle data sets of this size, although we encountered other problems. In discriminant analysis we estimate a linear or quadratic combination of explanatory variables to define a discriminant function which classifies each member of the sample into one of two groups. Discriminant analysis is a popular statistical method with specific underlying distributional assumptions (such as normality) used for classification when the data available are mostly continuous and when the populations are relatively balanced, something which is clearly not the case here. By contrast, Bayesian classifiers in general, and APRI in particular, make no distributional assumptions. APRI first selects key variables and then computes the prior probability distribution based on the training set for each selected variable. As new information or evidence arrives, we update the prior probability to compute a posterior. Therefore as we receive more information/evidence, we can form a more accurate profile of that particular customer or call.

---

[1] Uncollectible debt is the sum of NBD, direct adjustments, coin shortages and unbillable messages.

[2] We have selected our data segment carefully so that our data set proportions are more favorable for learning. For this paper we have an empirical uncollectible rate of 8-10%.

[3] We are using a software internally developed and implemented based on "CART" methodology.

Classification trees, when used for classification, are best when the feature set is mostly categorical and when, as with discriminant analysis, the population groups are relatively balanced. Note however that CART also has a "regression" component, but in our analysis we restricted ourselves to the classification tree side.

## 2 DATA AND ISSUES

The methodology which we describe is designed to address a large class of problems. Our motivation is largely practical as opposed to purely theoretical. We are faced with four major issues.

1) There are two populations; however, one has only a very low probability of being observed (~ 8-10% in our concentrated sample, < 2% in population) and accounts within the same population are not homogeneous; there may be many undefined sub-groups in each population.

2) The predictors are of mixed data types: some are categorical, some continuous. Their characteristics are:
i) The continuous variables can not be assumed to be normally distributed. ii) The categorical variables are themselves a mixed bag. Some have binary or trinary outcomes and could therefore be easily handled with dummy variables, others have many outcomes (e.g. 44, 130 or 1000+) and recoding them into binary outcome variables can be quite inefficient. Moreover, none of the categorical variables are ordered, precluding the use of a latent variable model.

Statistical discriminant analysis procedures for mixed variable cases do exist [see Krzanowski, 1980, Knoke, 1982, and Krzanowski, 1983]. However, they presume that none of the population groups is very small relative to the others (i.e. there are no "rare events"), and the continuous variables, conditioned on the outcome of a categorical variable, are normally distributed.

Clearly, both of these assumptions are violated by our problem. If not, then for many mixed variable cases, statistical distance metrics such as Mahalanobis distance are preferred to information theoretic metrics such as Kullback-Leibler distance (also known as relative entropy) [Krzanowski, 1983].[4] While classical discriminant analysis based on Mahalanobis distance metrics has been shown to be robust to departures from normality, our application brings with it too many violations of the classical assumptions. In preliminary analysis we have found discriminant analysis to perform poorly when compared to our new approach. The departure from normality together with the mixture of categorical and continuous data immediately suggests a non-parametric, or distribution-free approach.

3) For misclassification cost, the issue is very complex, not constant per class. Misclassification cost depends on the account treatment decision, the account itself, the customer's reaction, federal and state legal and regulatory constraints, and other factors.

Let us consider the issue of misclassification costs more carefully. In our telecommunications applications, it is highly undesirable from a business perspective to misclassify a good customer as bad. The costs associated with this event are different for every customer: it will depend on the expected lifespan of the customer with the company (a customer who would otherwise have remained with AT&T for two years is clearly more costly to lose than one who would have only remained for two months) and on the potential action taken with such a falsely identified customer (Do you shut him down? Do you call him? Do you send a letter?). The best option in fact could be "do nothing". This is a complex decision process with much uncertainty and would be best handled with the aid of NESDT, a decision theoretic normative expert system. Thus each account (or case) has a different misclassification cost associated with it, and the cost is in turn *conditional* on the action taken subsequent to identification and on the customer's reaction.

---

[4] All methods have the same first order asymptotic behavior. However, under the assumptions of normality, maximum likelihood achieves second order asymptotic superiority and discriminant analysis using Mahalanobis distance is optimal.

This is one of the key issues that led to the separation of fraud/uncollectible detection and account treatment. APRI focuses on detection, and passes the fraud/uncollectible probabilities to NESDT for account treatment recommendations. The expected misclassification cost for each account is implicitly computed in the NESDT using information such as the above factors, plus other factors (e.g., the expected account life, the account profitability, etc..). We believe it is improper to artificially bias the fraud/uncollectibe probabilities at the detection stage by imposing arbitrary misclassification cost.

4) Aside from the very nature of the data, its sheer size poses a problem as well. In our telecommunications network, a couple hundred million calls are placed daily. Each call provides us a couple hundred bytes of information. Hence we accumulate 40-50 giga-bytes (GB) of information daily. The data sets under analysis (a more detailed description is contained in Section 4) range anywhere from 8 mega-bytes (MB) to nearly 800 MB in size. Still, 800 MB is a tiny fraction of daily call detail information. In fact, since this application is in the telecommunication sector, any implementation is likely to impose similar data requirements on any modeling method. Hence the method, or more importantly the algorithm or software, must be able to cope with data sets of this magnitude. It is indeed a nontrivial accomplishment that APRI is able to handle this with relative ease.

# 3 CLASSIFICATION METHODS

Let us consider the case of two populations, $\pi_1$ and $\pi_2$, with a sample of $n_1$ and $n_2$ from each. We have two sets of characteristics: 1) Let Y be a c-length vector of continuous variables. 2) Let X be a q-length vector of discrete or categorical variables.

The classification problem can be described by the joint probability of classes/population $[\pi]$ and their attributes $[Y,X]$, i.e., $Pr\{\pi,Y,X\}$. The classification of the classes is based on the conditional probability of $Pr\{\pi|Y,X\}$. The various methods differ principally in the way that the conditional probability of the classes $\pi$ given the data is computed.

## 3.1 DISCRIMINANT ANALYSIS

Classical discriminant analysis assumes that the explanatory variables are jointly normally distributed, and that $Pr\{\pi|Y,X\}$ can be generated from a discriminant function. Under this assumption, if the covariance matrices of the two populations are equal, the optimal classification rule is the linear discriminant function. If we for the moment only consider the continuous variables Y, the sample linear discriminant function can be expressed as:

$$L(Y) = \{Y - \tfrac{1}{2}(\overline{Y}_1 + \overline{Y}_2)'\}S^{-1}(\overline{Y}_1 - \overline{Y}_2),$$

where $\overline{Y}_i$ are the sample mean vectors and S is the pooled sample covariance matrix. A new observation $Y_0$ is then classified into $\pi_2$ according to the rule:

$$L(Y_0) < c$$

where c is chosen according to objectives of the problem at hand. It turns out that the classification rule which minimizes the of misclassifications is obtained by letting

$$c = \log\left(\frac{n_2}{n_1}\right).$$

If the covariance matrices of the two populations are unequal, then we may use quadratic discriminant analysis which takes explicit account of the differing covariance matrices, $S_1$ and $S_2$. Again considering only the continuous variables for the moment, an observation Y would be classified into $\pi_1$ if

$$L(Y) = (Y - \overline{Y}_2)'S_2^{-1}(Y - \overline{Y}_2) - (Y - \overline{Y}_1)'S_1^{-1}(Y - \overline{Y}_1)$$

$$+ \log(|S_1|/|S_2|) > 2\log\left(\frac{n_2}{n_1}\right)$$

It comes as no surprise that, as the divergence between the two populations, measured by $|S_1|$ and $|S_2|$, decreases, the efficiency gain of using the quadratic instead of linear discriminant model decreases as well.

In nonparametric discriminant analysis, the conditional density of the class given the data is estimated directly, say via a kernel density algorithm. As the dimensionality of the feature set increases, the computational intensity increases dramatically. This is another manifestation of the well-known "curse of dimensionality", and in practice nonparametric discriminant analysis is useful for no more than three or four features or characteristics.

## 3.2 THE BAYESIAN NETWORK APPROACH

In theory, a "Bayesian Classifier" [Fukunaga, 1990] provides optimal classification performance, but in practice so far it has been considered impractical due to enormous data processing requirements. Recent advances in evidence propagation algorithms [Shachter 1989, Lauritzen 1988, Jenson 1990, Ezawa 1994] as well as advances in computer hardware allow us to explore this Bayesian classifier in the Bayesian network form [Cheeseman 1988, Herskovits 1990, Cooper 1992, Langley 1994, Provan 1995].

### 3.2.1 The Bayesian Network

To assess the conditional probability of $Pr\{\pi|Y,X\}$ of classes directly is often infeasible due to computational resource limitations. Therefore, we first assess the conditional probability of the attributes given the classes, $Pr\{Y,X|\pi\}$, and the unconditional probability of the classes, $Pr\{\pi\}$, using a preclassified training data set.

Figure 1 shows an example of a model where $Pr\{Y,X|\pi\}$ is further factorized to attribute level relationships $Pr\{Y_1|\pi\} * P\{Y_2|X_1,\pi\}*...* Pr\{X_q|Y_i,..,\pi\}$.

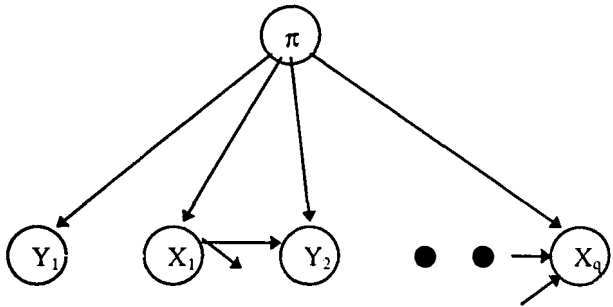

Figure 1: Bayesian Network Model

Once we assess the $Pr\{Y,X|\pi\}$, using Bayes rule, we compute the conditional probability $Pr\{\pi|Y,X\}$ with observed values of Y, X.

### 3.2.2 The association metric: entropy

The Advanced Pattern Recognition & Identification (APRI) system we developed is a Bayesian network-based supervised machine learning system resting on the above principle. We employ the entropy-based concept of mutual information to perform dependency selection. It first selects a set of variables and then selects a set of dependency among the selected variables using heuristically selected cumulative entropy thresholds. We settled on this approach to reduce the training time with special attention on repeated reading of training dataset as opposed to the methods used in [Herskovits 1990, Cooper 1992, Quinlan 1993, Provan 1995] where one variable or one segmentation of a variable is selected at a time.

Mutual information is defined for a pair of discrete random variables $X_i$, $X_j$ as

$$MI(X_i;X_j) = \sum_{outcomes} Pr(x_i,x_j) \log \frac{Pr(x_i,x_j)}{Pr(x_i)Pr(x_j)}.$$

Mutual information is used to rank population characteristics (or field nodes, to use the language of APRI) by how well they can discriminate between the two (or more) groups. It finds dependencies *between* these field nodes as well as to create a conditionally dependent model. Note that the mutual information metric is used only to determine model size and structure: how many field nodes to keep and which field to field node dependencies to allow for a conditionally dependent model. This is analogous in statistical regression analysis to variable selection for model specification.

Finding the optimal structure of a Bayesian Network is an NP hard problem, and we are therefore forced to engage in a heuristic search for the structure based on mutual information.

The first step of the training process for APRI is to create a schema file, which provides information how each variables are treated, feature selection information, variable definition (*e.g.*, continuous & and discrete, discretization methods or kernel density for continuous variables, etc.), cumulative entropy thresholds (percentage of total cumulative entropy), one for prime to field $T_{pf}$ ($0 \leq T_{pf} \leq 1$), the other for field to field $T_{ff}$ ($0 \leq T_{ff} \leq 1$), and moving window size (*i.e.*, whether it uses a single record or multiple records at a time), *etc.*.

*APRI Training Algorithm:*
*/ Input: schema file and training dataset */
1) a) if $\pi$ is continuous, compute class outcome set first
   b) compute outcome sets for $X_i \; \forall \; i = 1,...,q$ possible attribute variables, and $\pi$ (discrete classes).
2) a) compute all pairwise mutual information between prime and each field node $MI(\pi;X_i) \; \forall \; i = 1,...,q$ possible attribute variables
   b) select field variables in descending order until $T_{pf}$ is reached
3) a) compute all pairwise mutual information between field nodes conditional on the prime node $MI(X_j;X_i|\pi) \; \forall \; j \neq i$.
   b) select field to field dependencies in descending order until $T_{ff}$ is reached
4) compute $P\{X_i|C(X_i)\}$ and $P\{\pi\}$ where $C(X_i)$ represents parents of $X_i$ including $\pi$.

First (as a part of step 1), in a continuous classification task, APRI reads the training dataset to discretize or to estimate the kernel density of the classes. (This first step is unnecessary in discrete classification tasks.) Second (as the other part of step 1), it reads the training dataset to define the outcomes of each variable. It also discretizes continuous outcomes using entropy measures or estimates the kernel for example. On the third pass (step 2) it selects the variables to keep. On the fourth (step 3), it selects the dependencies among the previously selected variables. Finally (step 4), the dataset is read again to obtain the conditional probability distributions for the selected variables given the chosen dependencies. APRI performs better with higher $T_{pf}$ and $T_{ff}$ thresholds, but it is constrained by the computational environment (e.g., in the 32 bit operating system, a single model size is restricted to 2GB.)

We did not take the approach of [Herskovits 1990, Cooper 1992, Heckerman 1994, Provan 1995] to compute the joint probabilities $P\{B_s, D\}$, where $B_s$ is a Bayesian network structure, and D a dataset. First of all, our dataset violates most of the basic assumptions made for this approach; i.e, 1) it contains a mixture of categorical and continuous variables, 2) the records are not necessarily independent e.g., we have a sequence of calls from the same customer, and 3) we have many missing values. Secondly, the creation of an index tree[5] itself could be a problem. Just two variables with 1,000+ outcomes (e.g., originating and terminating cities) could create a link list of 1,000,000 cells. A few more variables like this makes it infeasible to construct and store the index tree, given the current computing environment. Thirdly, even if we succeed in creating the index tree[6], we will face a run time problem. In order to create a model, the training dataset needs to be read $n(u+1)$ times, where $n$ is a number of variables and $u$ is the maximum number of parents allowed. If you have a 4 million record training dataset with 33 variables (discussed later in Section 4.2), we would need to read the data 66 times with $u = 1$. In the case of discrete outcome classification, APRI needs to read the data set only four times regardless of the number of attributes or associated outcomes.

Lastly, even if we were to build an optimal Bayesian network model, it would become obsolete quite rapidly. The fraud/uncollectible debt problem environment is not static, but dynamic. In addition to the time related dynamics, more importantly, the detection model we create itself directly impacts the fraud/uncollectible problem environment of tomorrow. The better the model, the more quickly it becomes obsolete. The more successful we are in detecting and treating the fraud/uncollectible accounts or calls, the fewer accounts or calls will fit the existing model for future detection. The fraud/uncollectible problem simply evolves into a different shape and form and adjusts to the new environment. We may in fact not be able to find a "gold standard" Bayesian network structure. Therefore, given that the model will need to be retrained often, we put more emphasis on the ability to create a good model in a reasonable time using a very large dataset.

*Avoidance of probability of 1 ("Pruning"):* A particularly useful feature of APRI is its ability of avoiding the degeneracy of a probability at the tail end of the evidence propagation operation in the classification process. This feature allows us to separate the impact of the model's complexity from pure overfitting. The feature is simple. We skip the field node which causes $P\{\cdot\} = 1$ in the classification process as we would if we had missing information (or missing observation) for that field node. It is essentially the "pruning" of the tail-end attributes in the process of classification. Note that this effect of "pruning" is not explicitly observable on the network itself. It is implicitly hidden in the data structure under each field node.

*Moving windows:* APRI can model record dependency which is often manifested as the passage of time. It can create a model based on a set of variables over a specific length of a sequence. For example, the call detail dataset contains the time related information. APRI can exploit the sequence of calls to create a model. Others can do so only with extensive dataset reformatting[7].

### 3.3 OTHER METHODS: C4.5 AND CART

A well-known implementation of the entropy based metric is the machine learning program C4.5 by [Quinlan 1993]. In fact, C4.5 is a natural competitor to APRI because of this.[8] Similar to CART, C4.5 progressively splits the data set under investigation into subsets on increasing homogeneity. Quinlan noted that his entropy based gain criterion, used to determine such splits, exhibits a strong bias in favor of tests with many outcomes. He suggests scaling the gain criterion by a measure of marginal information gain from the split under consideration[9]. We are faced with a similar problem where nodes with many outcomes yield artificially high mutual information with respect to the prime node (classes) and other field nodes (under a conditionally dependent model). We are currently experimenting with several heuristic scaling or deflating factors.

C4.5 failed in building a model for our data set because of its sheer size. This is particularly disappointing because we have used only summary data for the test.

---

[5] See Cooper (1992), pp. 326-327.
[6] Since our dataset is large, to obtain a value in the index tree, it needs to access the dataset itself on a hard disk.
[7] When faced with a large dataset like ours, duplication and reformatting of the dataset should be minimized to reduce the training time and storage space.
[8] For an initial comparison to standard data sets, see [Ezawa and Schuermann, 1994].
[9] See Quinlan (1993), p.23.

The final business application will require call detail level data which will increase the data set size up to 100 fold. C4.5 in its present form is therefore not operationalizable for this application.

The classification and regression tree (CART) approach of [Breiman, Friedman, Olshen and Stone 1984] was originally developed because of the shortcomings of discriminant analysis when faced with mostly categorical data.[10] In theory CART accepts any distance or deviation metric, but the algorithm we used, based on the S software, employed the likelihood metric.[11] [Rao 1963] has provided results that demonstrate first-order asymptotic equivalence between models estimated by different metrics such as likelihood, entropy or minimum chi-square.

## 4 APPLICATION TO UNCOLLECTIBLE DETECTION

We now make a direct comparison of the performance of the three methods, where the benchmark is the "Do Nothing" strategy. At the very least, any method should be able to beat this alternative. We have concentrated the sample along a few dimensions for several reasons, mainly because from a business perspective the problem is interesting (at the moment) only for certain customer segments. In this paper we present the results of comparison using account summary data (40K - 80K records, 8 - 12 million bytes). Then we show preliminary results from APRI for out-of-sample prediction using detailed account activity dataset (4-6 million records, 600-800 million bytes of data).

For practical implementation, we need to train the model for a particular period and test it on a subsequent period of data. In this sense out-of sample prediction is not with a typical hold-out sample of the same period but a full set from a subsequent period. This makes the testing of the models harder but more realistic.

When considering goodness of fit, the conventional metric is the correct classification rate. Our concerns are somewhat different. We are principally worried about the number of customers classified as uncollectible, be they actually good or bad. Thus a method which classifies nearly everyone as collectible, yielding a respectable correct classification rate overall, will be useless for us; it is little different from the status quo of "Do Nothing".

In the Tables 1 and 2, the numbers in square brackets indicate the number of customer accounts for that cell. We include this information in addition to the classification percentages to highlight that even if the correct classification percentages are promising, the absolute numbers may not be simply because the populations are so unbalanced. Furthermore, beneath each table we have predicted collectible to uncollectible odds. This is useful because from an operational perspective, we are interested in just what proportion of the predicted population is in each group. Since the "Do Nothing" alternative makes no uncollectible predictions at all, we want to understand how any given model performs relative to this option, which if all models perform poorly, may indeed be the best option.

We focus on two aspects of results: 1) The ability to capture bad accounts, (i.e., if we cannot detect them, we cannot take actions.) 2) The ratio of falsely classified good accounts vs. correctly classified bad accounts (i.e., we don't want to be overwhelmed by false classification.) In other words, since there are many alternative treatment options available with various degree of positive to negative impact to the customers, a good model should capture large enough volume of bad accounts with reasonable volume of falsely identified good accounts.

### 4.1 PRELIMINARY ANALYSIS ON ACCOUNT SUMMARY DATA

The training sample sizes are 68,138 collectibles and 6,633 uncollectibles with 45 attributes yielding an unconditional probability of being uncollectible equal to 9.7%. Our final task is to predict/classify the dataset of period 2 based on a model created using the first period dataset since in the end, when this model is applied to real world business scenarios, it will be trained on one period and asked to perform in another later period. The testing sample sizes are 94,004 collectibles and 10,481 uncollectibles, yielding an unconditional probability of being uncollectible of 11.1%.

Model comparisons are presented in Table 1 below. For discriminant analysis we notice that a quadratic specification is clearly preferred over the linear one. This is simply telling us that the covariance matrices of the two populations are indeed quite different and pooling the observations to compute a joint covariance matrix is inefficient. In the end the linear model is not useful since it essentially classifies nearly everyone as collectible. The quadratic model manages to correctly classify about 20% of the bad accounts, but of the total pool predicted to be uncollectible, the false positives outnumber the true positives more than 5:1. In our application, false positives are very bad indeed if they are treated with the most severe action of denying access to the telecommunications network. One would never want to offend a good customer. Unconditionally in our sample, the probability of a random account being uncollectible is about 1 in 10. With the quadratic model we have improved this to 1 in 5. This is not good enough.

---

[10] See Breiman et al. (1984), pp. 15-16.
[11] Since the S version of CART is limited to small data sets, a nontrivial hurdle for this application, we made use of a modified version developed at Bell Laboratories. This C-based version can accept very large data sets and is otherwise identical to its S cousin.

Table 1: Model Comparison

| Model Type | F | C | V |
|---|---|---|---|
| Ideal | 0%<br>[0] | 100%<br>[10,481] | 0:1 |
| Linear | 0.25 %<br>[238] | 1.35%<br>[141] | 1.7:1 |
| Quadratic | 11.81 %<br>[11,106] | 20.78%<br>[2,178] | 5.1:1 |
| CART (pruned) | 0.09%<br>[86] | 0.69%<br>[75] | 1.1:1 |
| APRI ($\geq 50\%$*) | 6.89%<br>[6,668] | 25.91%<br>[2,808] | 2.4:1 |
| APRI ($\geq 70\%$*) | 2.34%<br>[2,269] | 12.45%<br>[1,349] | 1.7:1 |

F: False Classification of Good Accounts as Bad
C: Correct Classification of Bad Accounts
V: Volume Ratio -- F vs. C
*: Uncollectible(Bad Account) Probability Threshold

We found that the discriminant analysis results were very sensitive to minor changes in model specification. For categorical variables with many outcomes, grouping reduced the false classification rate, particularly the incidence of false positives. We believe this stems from two things: 1) The inefficient use of categorical information; to be precise, for large alphabet categorical variables, there will be many "empty cells", i.e. outcomes where no realization was observed in the sample. 2) The severe violation of the assumption of conditional normality; for example, linear and quadratic models yielded very different results.[12] This is especially distressing because it makes useful prediction wholly unreliable.

CART's pruning algorithm overprunes the tree with the result being that virtually the whole sample is classified in the majority class; the ability to detect bad accounts deteriorates substantially when compared with the unpruned tree (which had in excess of 4,000 nodes). In other words, with the pruned tree we hardly outperform the "Do Nothing" strategy. Though classification performance is generally considered to be better with the pruning than without it, most of the empirical experience is coming from relatively small data sets ( a few thousand records or less) with relatively balanced populations. It may not be prudent to assume this will hold true in very large datasets (a few hundred thousand, or several million records) with unbalanced populations. Pruning with unbalanced populations is very tricky and difficult indeed.

CART has difficulty dealing with unbalanced populations. Furthermore, the binary-split structure is quite limiting, and while interaction between variables is to some degree explicitly modeled, the notion of full conditional dependency as allowed by APRI is not possible.

We must keep in mind that this application is difficult partially for two reasons: 1) the populations are unbalanced; and 2) misclassification costs are highly uncertain, asymmetric, and vary from account to account. Thus any method which attempts to tackle this problem must address *both* issues. For instance, to solve the problem of unbalanced populations in CART by assigning asymmetric misclassification costs simply will not work when such asymmetric costs exist already.[13]

The most encouraging results were generated by APRI (with $T_{pf}$ = 95% and $T_{ff}$ = 35%.) At the 50% prediction level, i.e. the predicted probability of being uncollectible is $\geq$ 50%, APRI predicts 2.4 good customers for every bad. If we raise the predicted probability threshold to a more conservative 70%, we decline to a more palatable but still practically unacceptable 1.7 false to every true positive. Being more conservative means that we let slip through a few bad accounts, but the gain is falsely identifying far fewer good customers. We include this entry mostly for illustrative purposes, to show that for our application, a more conservative threshold may indeed be appropriate.

---

[12] We also estimated a logistic regression model with our data sets and found the results to be very similar to the linear discriminant model. In addition, we fit a nonparametric discriminant function using a kernel-based routine to estimate the conditional density. Its performance was similar to the linear model so we do not report it separately. It is interesting to note, however, that it took about 45 hours to estimate this model, versus about 3 minutes for the parametric versions, all on a SUN Sparc10 workstation.

[13] Addressing misclassification costs in machine learning is in fact an open area of research [Catlett 1995, Pazzani et al. 1994].

For the APRI model, the degree of accuracy depends on the degrees of conditional dependencies in the net. A conditionally independent model will be more robust, but it predicts very poorly.

While the account-level results show some promise, it is the call-detail information, a much richer data source, which should boost performance considerably. Below we show early results from call level detail information which indicate that the APRI model, performs even better than the other methods when these data are used.

### 4.2 PRELIMINARY ANALYSIS ON CALL LEVEL DETAIL DATA

So far we fit models only to summarized account level information. In actuality, our data is quite a bit richer than that: we have call level detail activity for every customer. We present preliminary results where we have fit a model using APRI (with $T_{pf}$ = 95% and $T_{ff}$ = 45%) of the call level detail information. The training dataset of period 1 consists of 4,014,721 records with 33 attributes (585 MB) and an unconditional probability of a call going bad of 9.90%. The period 2 dataset contains 5,351,834 records (773 MB) with an unconditional probability of a call going bad of 11.88%. It took about 4 hours to train the model and 5 hours to classify the period 2 dataset on a SUN Sparc 20 workstation. Table 2 shows the results of two uncollectible probability threshold setting. The results suggest that the information content in the call detail is indeed helpful in classification. In fact, when the uncollectible probability threshold is set to 0.7, we are able to correctly classify as many bad calls as incorrectly classify good calls while still capturing more than a fifth of all the bad calls.

Table 2: Performance with Call Detail

| Model Type | F | C | V |
|---|---|---|---|
| Ideal | 0 % [0] | 100% [635,611] | 0:1 |
| APRI ( ≥ 50%*) | 6.57% [309,784] | 31.86 % [202,500] | 1.5:1 |
| APRI ( ≥ 70%*) | 2.85 % [134,305] | 21.10% [134,131] | 1.0:1 |

F: False Classification of Good Accounts as Bad
C: Correct Classification of Bad Accounts
V: Volume Ratio -- F vs. C
*: Uncollectible(Bad Account) Probability Threshold

Figure 2 a) shows the changes in accuracy of correctly classified calls (good & bad), a standard metric for goodness of fit, correctly classified bad calls, and falsely classified good calls over the uncollectible probability threshold (from 10% to 90%). The higher the uncollectible probability threshold, i.e. the more conservative we are about classification,, the fewer good calls are falsely classified as bad, at the expense of the fewer bad calls being correctly classified.

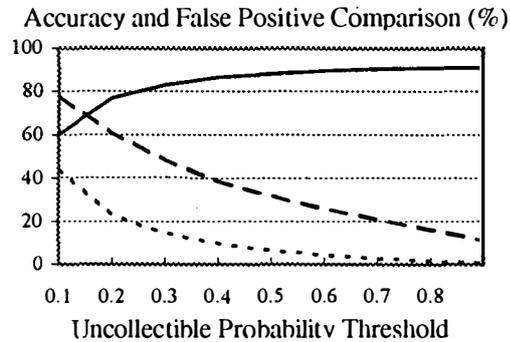

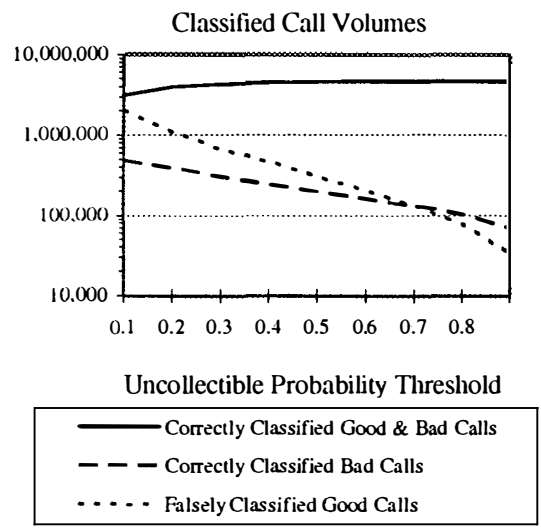

Figure 2: Accuracy & Call Volume vs. the Uncollectible Probability Threshold

If one were to focus only on accuracy in percentage, one would be quite happy with a lower, more aggressive predicted probability threshold since throughout the graph, the percentage of correctly classified good calls is larger than the percentage of falsely classified good calls. However, because of the disparity in volume of classes, simply looking at the classification percentages is not useful; we need to look at call volumes as the predicted probability threshold increases. The results are presented in Figure 2 b) where the Y-axis is in log scale. There we see that a more conservative approach is indeed warranted. Only at a threshold of around 0.7 do the correct positives outnumber the false positives in terms of volume.

## 5 AUTOMATED ACCOUNT TREATMENT RECOMMENDATIONS USING NORMATIVE EXPERT SYSTEM DEVELOPMENT TOOL

In section 4, we discussed the detection methods and found that even using the best method, the odds of classifying a good customer as uncollectible is still on the order of almost 1:1. Since good customers in this segment used in the analysis are some of the most valuable customers, we do not wish to take an action which may anger them and cause them to choose an alternative long distance company. We therefore need to handle customers very carefully. While determining the probability of uncollectible for each account is very important, we cannot base our actions on this probability alone. For example, we can imagine taking different actions based on a customer's profile. They will have different characteristics based on customer segment, and we want to treat them accordingly. If we do this properly with varying degrees of response, we can reduce the number of good customers' complaints as well as reducing the probability of losing them as customers. We can even increase the level of customer satisfaction by taking the right action.

In the telecommunication industry, where there are a few hundred million calls (several million bad calls) a day, to perform this task of detection and "rational" treatment in real-time with all these uncertainties is not humanly possible without assistance of a normative decision support system such as NESDT.

In descriptive research in psychology, it has been shown that people "systematically violate the principle of rational decision making when judging probabilities, making predictions or attempting to cope with probabilistic tasks. Biases in judgments of uncertainty events are often large and difficult to eliminate. The source of these biases can be traced to various heuristics or mental strategies that people use to process information" [Slovic, 1977]. The normative expert/decision support system is a formal tool that provides a means to overcome these inherent shortcomings and impose rationality when we face decision making under uncertainty [Heckerman 1990, Henrion 1992, Ezawa 1993].

We couple the APRI Bayesian network-based uncollectible model with NESDT for automated account treatment recommendation which is based on the influence diagram paradigm. An influence diagram is a generalized Bayesian network with decision and value nodes in addition to probabilistic nodes. By evaluating this diagram, the system recommends the optimal action given all available information and uncertainties. Preliminary testing of NESDT for automated account treatment recommendation is currently underway, and we plan to discuss the results in the coming conference.

## 6 SUMMARY

We have demonstrated the performance of Bayesian network based supervised machine learning system called APRI in the context of a rare binary outcome with mixed data types and compared it to more classical approaches such as discriminant analysis and CART. APRI's nonparametric approach lends itself well to this class of problems at the expense of being very data hungry. However, despite the fact that the outcome of interest occurs relatively rarely, the telecommunications sector is not exactly a data poor environment. In the comparison to other methods discriminant analysis proved to be the most robust to out-of-sample prediction, i.e. it suffered least from overfitting, while CART was the least robust. APRI came out ahead in overall performance and suffered only a little from overfitting the model to the training set.

When we set out to develop a model which would identify potentially fraudulent or uncollectible calls/customers, we were still ignorant of the important issue of falsely classifying good customers. In fact, in showing that traditional methods of thresholding or simple rule based methods and discriminant analysis are not good enough, and that the "Do Nothing" alternative outperforms them, we also demonstrated that merely beating this "Do Nothing" alternative, as the Bayesian network model does, is not sufficient. The handling of predicted uncollectible calls/customers in real-time under uncertainty is not feasible without intelligent system support, and so therefore we introduced a normative expert system based on NESDT to handle/recommend appropriate actions based on individual customer information under uncertainty. We believe that the combination and active interconnection of a Bayesian network model based on APRI for identification and intelligent system support based on NESDT for account treatment recommendation is a constructive and feasible tool for reducing exposure to net bad debt.

**Acknowledgements**

We wish to thank William Cohen, Steve Norton and two anonymous referees for helpful comments and suggestions.